\pdfoutput=1

\documentclass[11pt]{article}

\usepackage[preprint]{acl}

\usepackage{times}
\usepackage{latexsym}
\usepackage{array}
\usepackage{booktabs} 
\usepackage{multirow}
\usepackage{tabularx}
\usepackage{graphicx}
\usepackage{makecell}
\usepackage{caption}
\usepackage{subcaption}
\usepackage[titletoc]{appendix}
\usepackage{textcomp}
\usepackage{xspace}
\usepackage{soul} 
\usepackage{color}
\usepackage{xcolor}
\usepackage{amsmath}
\usepackage{amsfonts}
\usepackage{mathrsfs}
\usepackage{bm}
\definecolor{shadecolor}{rgb}{0.92,0.92,0.92}  
\usepackage{tcolorbox} 
\usepackage[export]{adjustbox}
\definecolor{mygray}{RGB}{230,230,240}
\definecolor{myblue}{RGB}{175, 238, 235}
\definecolor{deepgreen}{rgb}{0.0, 0.5, 0.0}
\usepackage[linesnumbered,ruled,vlined]{algorithm2e}
\usepackage{multicol}
\usepackage{tikz}
\usetikzlibrary{shapes, positioning}
\usepackage{setspace}
\usepackage{hyperref}
\usepackage{placeins}

\usepackage{enumitem}
\usepackage[skip=5pt]{caption}
\usepackage{xcolor} 

\usepackage[T1]{fontenc}

\usepackage[utf8]{inputenc}

\usepackage{microtype}

\usepackage{inconsolata}

\usepackage{graphicx}
\usepackage{hyperref}
\title{A Reasoner for Real-World Event Detection: Scaling Reinforcement Learning via Adaptive Perplexity-Aware Sampling Strategy}

\author{
\textbf{Xiaoyun Zhang}$^{1,\dagger}$\quad
\textbf{Jingqing Ruan}$^{1,\dagger, \ast}$\quad
\textbf{Xing Ma}$^{1,\dagger}$\quad
\textbf{Yawen Zhu}$^{1}$\\
\textbf{Jiansong Chen}$^{1, \ast}$\quad
\textbf{Ke Zeng}$^{1}$\quad
\textbf{Xunliang Cai}$^{1}$\\
$^1$Meituan\\
\texttt{zhangxiaoyun15@meituan.com} \quad \texttt{ruanjingqing@meituan.com}
}

\begin{document}
\maketitle

\renewcommand{\thefootnote}{\fnsymbol{footnote}}
\footnotetext[2]{Xiaoyun Zhang, Jingqing Ruan, and Xing Ma contributed equally to this work.}
\footnotetext[1]{Corresponding author.}
\renewcommand{\thefootnote}{\arabic{footnote}}

\begin{abstract}

Detecting abnormal events in real-world customer service dialogues is highly challenging due to the complexity of business data and the dynamic nature of customer interactions. 
Moreover, models must demonstrate strong out-of-domain (OOD) generalization to enable rapid adaptation across different business scenarios and maximize commercial value.
In this work, we propose a novel Adaptive Perplexity-Aware Reinforcement Learning (APARL) framework that leverages the advanced reasoning capabilities of large language models for abnormal event detection. 
APARL introduces a dual-loop dynamic curriculum learning architecture, enabling the model to progressively focus on more challenging samples as its proficiency increases. This design effectively addresses performance bottlenecks and significantly enhances OOD transferability.
Extensive evaluations on food delivery dialogue tasks show that our model achieves significantly enhanced adaptability and robustness, attaining the highest F1 score with an average improvement of 17.19\%, and an average improvement of 9.59\% in OOD transfer tests. This method provides a superior solution for industrial deployment of anomaly detection models, contributing to improved operational efficiency and commercial benefits.

\end{abstract}

\section{Introduction}



In the domain of real-world customer service~\citep{naik2024leveraging,yang2024optimization,ngai2021intelligent,zou2021topic}, effectively detecting abnormal events in conversations among users, merchants, and service agents is crucial for timely issue resolution and proactive risk mitigation.
For example, in food delivery scenarios~\citep{gao2021deep}, when a user repeatedly complains about delayed deliveries, it is essential to analyze the context of the conversation to identify the root cause of the issue, such as slow food preparation by the merchant or the courier failing to pick up the order. 
Providing better solutions based on this analysis can enhance user satisfaction, helping the platform retain more customers and thereby boosting its competitive edge in the industry.

Existing methodologies for abnormal event detection face two predominant limitations.
Traditional approaches relying on specialized small models (e.g., BERT-based classifiers)~\citep{devlin2019bert,chen2024bge,li2023gte,qasim2022fine,prabhu2021multi,yu2019improving} necessitate training and maintaining multiple task-specific models, resulting in prohibitive annotation costs (averaging 120 hours per event) and poor scalability for emerging event types. While large language models (LLMs) address these issues through few-shot prompting~\citep{touvron2023llama,kaplan2020scaling,min2022rethinking,wei2022chain}, their performance exhibits unacceptable sensitivity to prompt engineering and does not perform well in vertical domains such as the food delivery sector.
Furthermore, supervised fine-tuned LLMs~\citep{wei2021finetuned,openai2023gpt4,ouyang2022training,wake2023chatgpt,sun2023fine} can adapt to some vertical scenarios, which have been widely adopted in various industrial fields~\citep{dai2024does,ravikumar2025study}.
However, it tends to the memorization bias~\citep{chu2025sft}, where models overfit to training data, limiting their generalization to unseen data. This memorization suggests that performance gains may stem more from learned patterns rather than deep task understanding.

\begin{figure}[htbp!]
  \centering
  \includegraphics[width=1.03\linewidth]{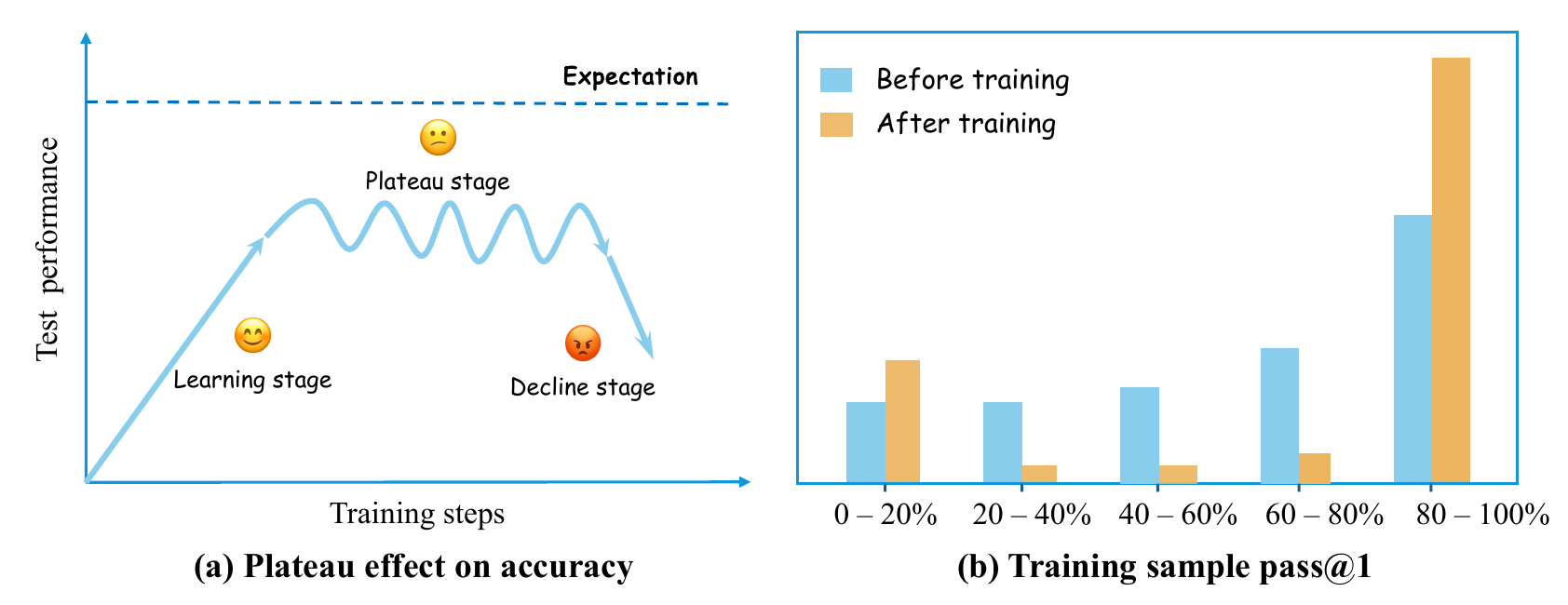}  
  \caption{(a) Training Plateau effect on accuracy. (b) Sample pass@1 distribution changes with conventional RL training.}
  \label{fig:intro}
\end{figure}

To overcome these bottlenecks, we explore the potential of reasoning capabilities in large models to enhance problem-solving in customer service dialogue scenarios. Under a post-training framework with reinforcement learning (RL), reasoning-focused large models~\citep{jaech2024openai,guo2025deepseek,team2025kimi,gemeni_thinking} such as OpenAI's o1/o3~\citep{jaech2024openai} and DeepSeek-R1~\citep{guo2025deepseek} demonstrate exceptional capabilities utilizing chain-of-thought reasoning. Despite RL's strong theoretical performance, its real-world applications in customer service dialogues face significant implementation challenges. Our preliminary experiments and existing research~\citep{arrieta2025o3,mondillo2025comparative,evstafev2025token} indicate that training challenges in domain-specific tasks primarily include: (1) a plateau effect on accuracy, where early convergence is followed by stagnation or decline~\citep{schaul2019ray}, as shown in Figure~\ref{fig:intro}(a); (2) a bimodal distribution in sample pass@1 results after RL training, indicating difficulties learning from hard samples~\citep{slade1993bimodal}, as demonstrated in Figure~\ref{fig:intro}(b).

Based on these observations, we propose a novel Adaptive Perplexity-Aware Reinforcement Learning (APARL) framework to address these challenges through a dual-loop architecture. In the outer loop, APARL employs an adaptive perplexity-aware sampling strategy to dynamically select samples based on complexity; the inner loop utilizes a rule-guided reinforcement learning mechanism to optimize model learning, encouraging exploration of diverse reasoning strategies without requiring explicit CoT annotations. This dual-loop strategy enables dynamic curriculum learning, allowing the model to gradually focus and explore more challenging samples as its capabilities improve, enhancing sample utilization efficiency and performance optimization limits. 






Extensive experiments and evaluations on food delivery dialogue tasks demonstrate that APARL effectively enhances model accuracy and robustness, achieving the highest F1 score with an average increase of 17.19\% across all benchmarks; in OOD task testing, the average increase was 9.59\%. Our method provides a superior solution for the industrial application of abnormal detection models, aiding in the improvement of business operational efficiency and economic benefits.

\section{Problem Statement}

\begin{figure}[htbp!]
  \centering
  \includegraphics[width=0.85\linewidth]{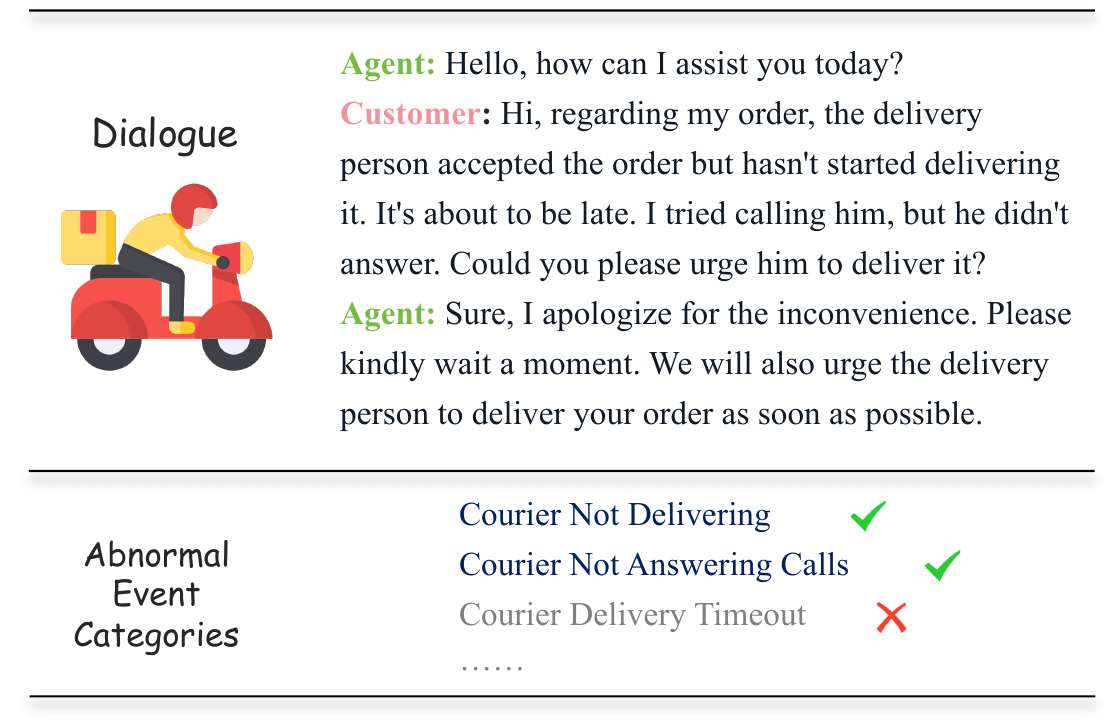}  
  \caption{Abnormal event detection in the dialogue.}
  \label{fig:problem_des}
\end{figure}

In this section, we formalize the abnormal event detection in customer service dialogues as a decision-making process~\cite{howard1960dynamic,baxter1995markov}.
Let a customer service dialogue be a sequence of $T$ utterances $D = \{u_1, ..., u_T\}$ where $u_t \in \mathcal{U}$ represents the $t$-th utterance (user/agent/  merchant). 
As illustrated in Figure~\ref{fig:problem_des}, our task is to analyze the dialogue and give the names of abnormal events.
We formulate abnormal event detection as RL problem with:
(1) \textbf{State space}: $\mathcal{S} = \{ s_t = (e_p, u_{t-k:t}) \}$, where $p$ denotes the event descriptions and $u_{t-k:t}$ denotes dialogue window of $k$ recent utterances; (2) \textbf{Action space}: $\mathcal{A} = \{a_1, ..., a_M\} $, where $M=20$ predefined abnormal event types (e.g., merchant\_delay, courier\_issue); (3) \textbf{Reward function}: $r$ contains the accuracy reward and the format penalty.


\section{Methodology}

\subsection{Overview}

\begin{figure*}
    \centering
    \includegraphics[width=0.91\linewidth]{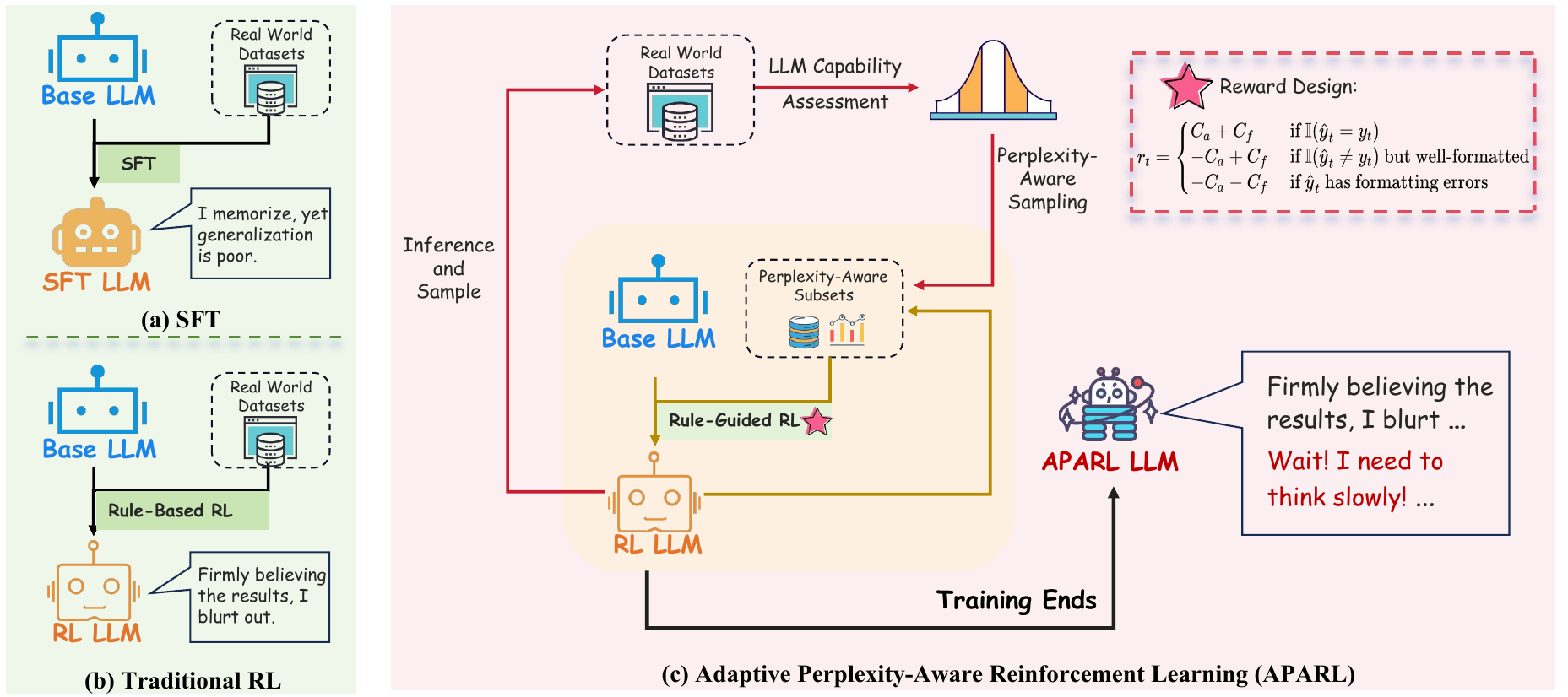}
    \caption{(a) SFT, (b) traditional RL, and (c) overview of our proposed APARL: integrating rule-guided RL in the inner loop and the adaptive perplexity-aware sampling strategy in the outer loop to improve the SFT and traditional RL issues.}
    \label{fig:mian_framework}
\end{figure*}

In order to fully leverage the advantages of reasoning models in detecting real-world abnormal events, we explores the training methods for model reasoning and addressing issues in customer service dialogue scenarios within the RLVR framework. However, preliminary experimental results indicate that the model's training performance has not met the expected standards. Through multiple rounds of experimental observations, the following phenomena were noted: (1) Accuracy converges early in training, followed by a plateau effect, and eventually shows a declining trend~\citep{schaul2019ray}, as illustrated in Figure~\ref{fig:intro}(a); (2) There are discrepancies compared to the experimental results reported in existing studies, with a decreasing trend in the length of reasoning chains during training; (3) The pass@1 results of samples post-reinforcement learning training exhibit a bimodal distribution characteristic~\citep{slade1993bimodal}, with a slight increase in the number of difficult problems, as shown in Figure~\ref{fig:intro}(b).

The fundamental cause of the observed training trends lies in the fact that, under the same sampling budget constraints, some intermittently solvable problems may not be sampled to the correct answer, while some simple problems with intermittent errors have correct answers and provide effective gradient signals. The model gradually converges towards the correct response trajectory of simple samples, leading to a degradation in the solving ability of complex problems that originally had intermittent solvability. To address the challenge of performance stagnation in reinforcement learning training, we propose the Adaptive Perplexity-Aware Reinforcement Learning (APARL) framework, as shown in Figure~\ref{fig:mian_framework}, which includes a dual-loop architecture:

\begin{itemize}[leftmargin=10pt]

\item \textbf{\textit{Outer Adaptive Sample Strategy}}: Automatically optimizes training data composition using the perplexity based on the model’s average proficiency level and training data distribution.

\item \textbf{\textit{Inner Rule-Guided RL}}: Activates foundational reasoning through structured reward signals aligned with domain expertise.

\end{itemize}

\subsection{Outer loop: Adaptive Perplexity-Aware Sampling Strategy}

To address performance stagnation and overfitting to simple samples, the outer loop of our APARL framework dynamically adjusts the sampling probability of each training sample according to the model’s capability at every training step. This ensures that the model is consistently exposed to appropriately challenging samples.

\textbf{LLM Capability Assessment.} At each training step, we perform Monte Carlo sampling on the current LLM policy $\pi_{\theta}$, conducting $k$ independent generations for each prompt $x_i$ in the current mini-batch $B=\{x_i\}_{i=1}^b$. We define the model's proficiency level $\mu_p \in [0,1]$ through:

\begin{equation}
p_i = \frac{1}{k}\sum_{j=1}^k \mathbb{I}(f_\pi(x_i^{(j)})=y_i), \quad  \mu_p^{(t)} = \frac{1}{N}\sum_{i=1}^N p_i,
\end{equation}
where $p_i$ measures empirical success rate, $b$ is the batch size, $t$ denotes training iteration, and $f_\pi$ represents the policy function.

\textbf{Perplexity-Aware Sampling.} 
Building upon the batch-wise capability assessment, we introduce a power-function-based perplexity-aware sampling probability to directly determine whether each sample $x_i$ in the mini-batch is selected for training at the current step. Specifically, for each sample, we compute its selection probability as follows:
\begin{equation}
P(x_i) = \left(\frac{p_i}{\mu_p + \epsilon}\right)^{t\mu_p} \cdot \left(\frac{1-p_i}{1-\mu_p + \epsilon}\right)^{t(1-\mu_p)}
\label{eq:sample}
\end{equation}
where $p_i$ is the empirical success rate of the model on $x_i$, $\mu_p$ is the average proficiency over the current batch, $t > 0$ is a sharpness parameter controlßling the concentration of the distribution, and $\epsilon$ is a small constant (e.g., $1\times10^{-8}$) added to the denominators for numerical stability.

Similar to DAPO~\citep{yu2025dapo}, the sampling probability computed by Eq~\eqref{eq:sample} naturally becomes zero for samples where $p_i$ equals 0 or 1, thereby eliminating the influence of ineffective gradient samples on the learning process. Additionally, the peak of the sampling function is located at $\mu_p$, meaning that samples whose difficulty is close to the current model’s average proficiency are more likely to be selected (i.e., their sampling probability approaches 1). After applying the sampling filter, if the number of selected samples is insufficient to fill the training batch size, the same sampling procedure will be performed on the subsequent batch to supplement the batch. This design exhibits two essential properties:

\begin{itemize}[leftmargin=15pt]

    \item \textit{Capability-driven Focus}: In the early stage of training, when $\mu_p$ is small, the distribution peak is close to the low-difficulty region, so the model tends to sample easier examples, facilitating rapid skill acquisition.

    \item \textit{Challenge Emphasis}: As training progresses and $\mu_p$ increases, the distribution peak shifts toward higher difficulty, and the model naturally focuses on harder or borderline samples, shifting the training emphasis to more challenging tasks.

\end{itemize}

The perplexity-aware data dynamically guides the model toward deeper cognitive engagement, avoiding stagnation in simplistic problem-solving patterns. This process mimics curriculum learning, where the training focus gradually shifts to harder tasks as the model improves, ultimately converging to an optimal policy through synergy with the inner loop.

\subsection{Inner loop: Rule-Guided RL}

In abnormal event detection tasks, where sample labels exhibit high specificity, rule-guided RL demonstrates significant application advantages. This study integrates rule-guided RL with the emergent properties of Chain-of-Thought (CoT) reasoning, leveraging the synergistic effects of test-time scaling benefits and RL scaling effects to construct a scalable training framework.

\textbf{RL Algorithm.} We employ a modified DAPO~\citep{yu2025dapo} algorithm as the core RL module. Preliminary ablation studies demonstrate that omitting the KL divergence constraint leads to significant instability during training. To mitigate this, we reintroduce the KL divergence term. Specifically, we retain both the token-level policy gradient loss computation and the Clip-Higher strategy. In addition, the KL divergence constraint is decoupled from the advantage estimation procedure and is instead incorporated as an explicit regularization term within the loss function, following the optimization practices recommended by DeepSeek-Math~\citep{Shao24:Deepseekmath}. The revised overall loss function is formally defined in Eq~\eqref{eq:rl}:

\begin{align}
\mathcal{L}(\theta) = -\mathbb{E} &\Bigg[ 
    \frac{1}{G} \frac{1}{|o_i|}\sum_{i=1}^G  \sum_{t=1}^{|o_i|} 
    \Big( 
    \min \big( r_{i, t} \hat{A}_{t}, \nonumber \\
    &\mathrm{clip}(r_{i, t}, 1\!-\!\varepsilon_{low}, 1\!+\!\varepsilon_{high}) \hat{A}_{t} \big) \nonumber \\
    &- \lambda \mathbb{D}_{\mathrm{KL}}\left[\pi_\theta \| \pi_{\mathrm{ref}}\right]
    \Big)
\Bigg]
\label{eq:rl}
\end{align}
where the advantage function $\hat{A}_t$ and the importance sampling coefficient $r_{i, t}$ are computed as
\begin{equation}
\hat{A}_t = \frac{r_t-\mu_r}{\sigma_r}, \quad r_{i, t} = \frac{\pi_\theta\left(o_{i, t} \mid q, o_{i,<t}\right)}{\pi_{old} \left(o_{i, t} \mid q, o_{i,<t}\right)}
\end{equation}

Additionally, we adopt the unbiased KL estimation as recommended in GRPO algorithms to enhance stability, as shown in Eq.~\eqref{eq:kl}:

\begin{equation}
\begin{split}
\mathbb{D}_{\mathrm{KL}}\left[\pi_\theta \| \pi_{\mathrm{ref}}\right]=\frac{\pi_{\mathrm{ref}}\left(o_{i, t} \mid q, o_{i,<t}\right)}{\pi_\theta\left(o_{i, t} \mid q, o_{i,<t}\right)} - \\
\log \frac{\pi_{\mathrm{ref}}\left(o_{i, t} \mid q, o_{i,<t}\right)}{\pi_\theta\left(o_{i, t} \mid q, o_{i,<t}\right)}-1.
    \label{eq:kl}
\end{split}
\end{equation}


\textbf{Reward Design.} For abnormal event detection, the rule-based reward function includes a format reward and an answer accuracy reward. Following GRPO-LEAD~\citep{zhang2025grpolead} and RM-R1~\citep{chen2025rmr1}, we assign negative scores to responses that fail to meet format or correctness requirements. Both components are initialized with negative constants ($-C$, $C>0$) and switched to positive values if the output is correct or well-formatted. The final reward is computed as the sum of these and serves as the advantage estimator for policy gradient optimization. The reward function is formalized as Eq~\eqref{eq:reward}:
\begin{equation}
r_{t} = \begin{cases} C_a + C_f & \text { if $\mathbb{I}(\hat{y}_t = y_t)$} \\ -C_a + C_f & \text { if $\mathbb{I}(\hat{y}_t \neq y_t)$ but well-formatted } \\ -C_a - C_f & \text { if $\hat{y}_t$ has formatting errors. }\end{cases}
\label{eq:reward}
\end{equation}

\section{Experiment}


\subsection{Experimental Setting}

\textbf{Dataset Selection}: 
We benchmark our proposed framework against baseline models using a comprehensive industrial food delivery dataset. This dataset, sourced from Meituan's online logs, comprises 55,000 training samples and 9,000 testing samples. It features multi-turn conversations and is structured to classify 20 distinct abnormal events. The classification targets are meticulously annotated by our human customer service team.

\textbf{Evaluation Metrics}: We validate our method using precision, recall, and F1-score metrics on the 9,000 test samples to ensure robust performance assessment.

\textbf{Model Configuration}: Our experiments utilize Qwen-14B-Instruct and DeepSeek-R1-Distill-Qwen-14B as the base models for RL training. This configuration is selected to thoroughly evaluate the applicability of both reasoning and standard models with our APARL, thereby demonstrating the generalizability of our approach.

\textbf{Comparison Methods}: We employ the following comparisons: (1) Trained small models: BGE-M3\footnote{https://huggingface.co/BAAI/bge-m3}; (2) API calls using GPT-4o/DeepSeek-V3 (Instruct model) and o1-preview/DeepSeek-R1 (Reasoning model); (3) Supervised Fine-Tuning (SFT) on the aforementioned base models; and (4) the GRPO and DAPO RL methods versus APARL.

\subsection{Main Results}

\begin{table}[ht]
\caption{Performance Comparison with Different Strategies}
\label{tab:structured_comparison}
\centering
\resizebox{0.48\textwidth}{!}{ 
\begin{tabular}{llccc}
\toprule
\multirow{2}{*}{\textbf{Category}} & 
\multirow{2}{*}{\textbf{Model}} & 
\multicolumn{3}{c}{\textbf{Performance}} \\
\cmidrule(lr){3-5}
 & & Precision & Recall & F1 \\
\midrule
\multirow{1}{*}{\textbf{Small Models}} 
 & BGE-M3 & 78.53\% & 69.90\% & 73.96\% \\
\midrule
\multirow{4}{*}{\textbf{API Models}} 
 & GPT-4o & 72.52\% & 77.40\% & 74.88\% \\
 & o1-preview & 65.43\% & 72.35\% & 68.72\% \\ 
 & DeepSeek-V3 & 74.76\% & 68.56\% & 71.53\% \\
 & DeepSeek-R1 & 79.85\% & 69.59\% & 74.37\% \\
\midrule
\multirow{2}{*}{\textbf{Base Models}}
 & \makecell[l]{DeepSeek-R1-Distill-Qwen-14B} & 62.58\% & 76.96\% & 69.03\% \\
 & Qwen-14B-Instruct & 73.63\% & 56.09\% & 63.67\% \\
\midrule
\multirow{2}{*}{\textbf{Ours}} 
 & \makecell[l]{DeepSeek-R1-Distill-Qwen-14B+APARL} & \textbf{82.67\%} & \underline{81.90\%} & \underline{82.28\%} \\
& \makecell[l]{Qwen-14B-Instruct+APARL} & \underline{80.38\%} & \textbf{86.61\%} & \textbf{83.38\%} \\
\bottomrule
\end{tabular}
}
\end{table}

The results presented in Table~\ref{tab:structured_comparison} highlight the significant improvements achieved by our APARL across various models and metrics. APARL \textbf{outperforms leading API-based models}, such as \textit{DeepSeek-R1}, by attaining a higher F1 score (83.38\% vs. 74.37\%). This indicates the practical applicability of APARL, which can save more costs.
Moreover, the results also reveal that APARL \textbf{surpasses the performance of traditional small models} like \textit{BGE-M3} (73.96\% F1), further validating the advantages of our RL-driven approach in abnormal event detection tasks. Moreover, our strategy, when \textbf{applied to both instruct and distill models, demonstrates superior performance} in terms of F1 scores. Specifically, \textit{DeepSeek-R1-Distill-Qwen-14B+Ours} achieves an F1 score of 82.28\% while \textit{Qwen-14B-Instruct+Ours} reaches 83.38\%.


Another noteworthy finding is that the reasoning patterns acquired by \textit{DeepSeek-R1-Distill-Qwen-14B} from DeepSeek-R1 \textbf{did not yield significant improvements in business-oriented evaluations}. In fact, the distilled model performed even worse compared to Qwen-14B-Instruct after RL training. Our experimental analysis may provide additional insights for the industry when selecting models.

Overall, APARL achieved a 17.19\% average improvement over all the baselines. Our training method effectively addresses three real-world challenges: (1) Enhanced recall improves long-tail coverage, allowing detection of unusual issues like unique customer complaints. (2) It maintains high precision while identifying more issues, ensuring compliance with business rules like privacy policies without losing accuracy. (3) Consistent improvements across various AI models demonstrate the method's versatility, proving it's not limited to specific technical setups.

\subsection{Ablation Study}

\begin{table}[ht]
\caption{Ablation on Different Data/Training Strategies}
\label{tab:abl_study}
\centering
\resizebox{0.48\textwidth}{!}{ 
\begin{tabular}{llccc}
\toprule
\multirow{2}{*}{\textbf{Model}} & 
\multirow{2}{*}{\textbf{Method}} & 
\multicolumn{3}{c}{\textbf{Performance}} \\
\cmidrule(lr){3-5}
 & & Precision & Recall & F1 \\
\midrule

\multirow{5}{*}{\textbf{DeepSeek-R1-Distill-Qwen-14B}}
 & \makecell[l]{Base model} & 62.58\% & 76.96\% & 69.03\% \\
 & \makecell[l]{+SFT } & 74.28\% & 79.84\% & 76.96\% \\
 & \makecell[l]{+GRPO } & 67.64\% & \textbf{92.06\%} & 77.98\% \\
 & \makecell[l]{+DAPO } & \underline{78.87\%} & 78.95\% & \underline{78.91\%} \\
 & \makecell[l]{+Our APARL} & \textbf{82.67\%} & \underline{81.90\%} & \textbf{82.28\%} \\
\midrule
\multirow{5}{*}{\textbf{Qwen-14B-Instruct}}
 & Base model & 73.63\% & 56.09\% & 63.67\% \\
 & \makecell[l]{+SFT } & \textbf{83.05\%} & 66.12\% & 73.63\% \\
 & \makecell[l]{+GRPO } & 77.63\% & 74.55\% & 76.06\% \\
 & \makecell[l]{+DAPO } & 76.88\% & \underline{80.15\%} & \underline{78.48\%} \\
 & \makecell[l]{+Our APARL} & \underline{80.38\%} & \textbf{86.61\%} & \textbf{83.38\%} \\

\bottomrule
\end{tabular}
}
\end{table}

The ablation study was conducted using two base models: \textit{DeepSeek-R1-Distill-Qwen-14B} and \textit{Qwen-14B-Instruct}. We evaluated the following configurations: the base model, SFT with the full dataset, conventional RL(GRPO and DAPO), and APARL.

For \textit{DeepSeek-R1-Distill-Qwen-14B/Qwen-14B-Instruct}, applying APARL resulted in an F1 score of 82.28\%/83.38\%, which is a notable improvement over both the full SFT(6.91\%/13.24\%).
This indicates that our dynamic approach effectively balances precision and recall, outperforming two conventional RL methods by 4.89\%/7.93\% in the F1 score.
These results show the potential and advantages of our method for industrial applications.

\subsection{OOD Testing}

\begin{table}[ht]
\caption{OOD Testing on Different Data/Training Strategies}
\label{tab:ood_study}
\centering
\resizebox{0.48\textwidth}{!}{ 
\begin{tabular}{llccc}
\toprule
\multirow{2}{*}{\textbf{Model}} & 
\multirow{2}{*}{\textbf{Method}} & 
\multicolumn{3}{c}{\textbf{Avg. Performance}} \\
\cmidrule(lr){3-5}
 & & Precision & Recall & F1 \\
\midrule

\multirow{1}{*}{\textbf{Small Models}}
 & \makecell[l]{BGE-M3} & / & / & / \\
\midrule

\multirow{3}{*}{\textbf{API Models}}
 & \makecell[l]{GPT-4o} & 71.81\% & 70.22\% & 71.00\% \\
 & \makecell[l]{DeepSeek-V3} & 66.78\% & 71.69\% & 69.16\% \\
 & \makecell[l]{DeepSeek-R1} & 70.43\% & 72.40\% & 71.26\% \\
\midrule

\multirow{5}{*}{\textbf{DeepSeek-R1-Distill-Qwen-14B}}
 & \makecell[l]{Base model} & 54.69\% & \textbf{87.75\%} & 67.36\% \\
 & \makecell[l]{+SFT } & 66.13\% & 78.03\% & 71.53\% \\
 & \makecell[l]{+GRPO } & 68.30\% & 82.78\% & 74.82\% \\
 & \makecell[l]{+DAPO } & 76.13\% & 73.66\% & 74.84\% \\
 & \makecell[l]{+Our APARL} & \textbf{78.99\%} & 79.11\% & \textbf{79.01\%} \\
\midrule
\multirow{5}{*}{\textbf{Qwen-14B-Instruct}}
 & Base model & 56.82\% & \underline{86.19\%} & 68.46\% \\
 & \makecell[l]{+SFT } & 66.54\% & 79.16\% & 72.26\% \\
 & \makecell[l]{+GRPO } & \underline{78.07\%} & 71.33\% & 74.53\% \\
 & \makecell[l]{+DAPO } & 77.13\% & 77.51\% & 77.30\% \\
 & \makecell[l]{+Our APARL} & 77.45\% & 79.92\% & \underline{78.65\%} \\

\bottomrule
\end{tabular}
}
\end{table}

Table~\ref{tab:ood_study} summarizes the out-of-distribution (OOD) testing results across different model configurations and training strategies. We report the average performance metrics (Precision, Recall, F1) over three anonymized business domains(to ensure business confidentiality), whose respective test set sizes are 8.8k, 8.5k, and 5k. It should be noted that the BGE-M3 model cannot be directly transferred for use, as it has a fixed classification head. More experiments details for each business can be found in Appendix \ref{app:results}.

From the results, we observe that our proposed APARL method consistently outperforms all baselines. For \textit{DeepSeek-R1-Distill-Qwen-14B}, APARL achieves the highest F1 score (79.01\%), representing a 7.48\% absolute improvement over SFT and outperforming both GRPO and DAPO. Similarly, for \textit{Qwen-14B-Instruct}, APARL yields an F1 of 78.65\%, surpassing both SFT and other RL methods. Comparing with strong API-based models (e.g., GPT-4o, DeepSeek-V3), our APARL-trained models demonstrate highly competitive OOD generalization, with F1 scores exceeding those of GPT-4o and DeepSeek-R1 by 8.01\% and 7.75\%, respectively, on average. 

Overall, APARL achieved a 9.59\% average improvement over all the baselines. This highlights the robustness of our approach under distributional shifts, which is crucial for real-world industrial deployment.


\subsection{Dynamics on Validation and Training}

\begin{figure}[htbp]
    \centering
    \begin{subfigure}{0.235\textwidth}
        \centering
        \includegraphics[width=\linewidth]{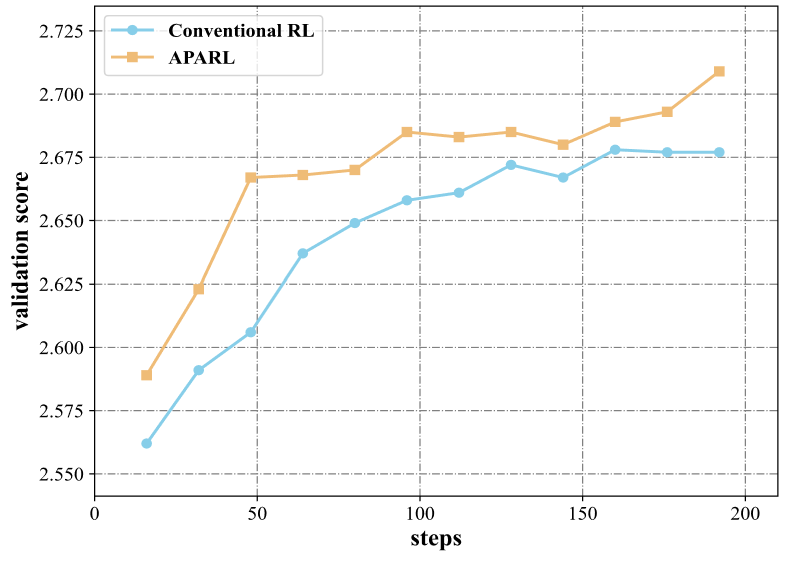}
        \caption{Validation Score}
        \label{fig:reward1}
    \end{subfigure}
    \begin{subfigure}{0.235\textwidth}
        \centering
        \includegraphics[width=\linewidth]{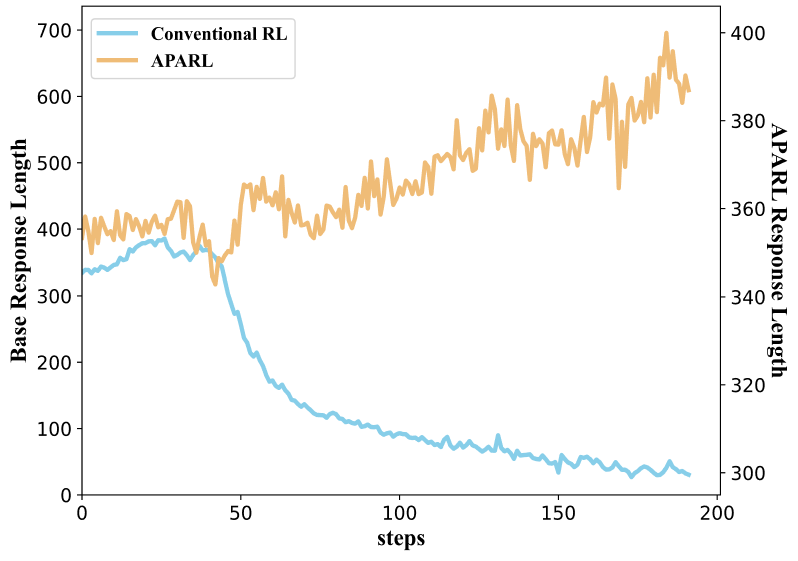}
        \caption{Response Length}
        \label{fig:reward2}
    \end{subfigure}
    \caption{Visualization on Dynamics.}
    \label{fig:overall}
\end{figure}

The validation and training dynamics depicted in Figure~\ref{fig:overall} highlight the effectiveness of our adaptive strategy. In the left panel, the reward trajectory of the adaptive strategy exhibits a more rapid upward trajectory and ultimately converges to a higher reward region. This characteristic stems from the dynamic adjustment of sample difficulty, enabling the model to effectively consolidate learned strategies at each stage while progressively enhancing its capabilities. The right panel simultaneously demonstrates the variation in response length. Compared to baseline methods, our strategy achieves significantly longer response lengths, as the model gradually increases its exploration of complex scenarios. This adaptive adjustment stimulates the model's capacity for deep reasoning and enables robust, context-aware decision-making in reinforcement learning tasks.


\section{Conclusion}


In conclusion, we are the first to apply reasoning models to abnormal event monitoring in customer service workflows. We propose a novel reinforcement learning approach that addresses real-world challenges by integrating domain-specific rewards and adaptive perplexity-aware sampling. Our framework significantly boosts abnormal event detection and resolution.
Extensive experiments on a food delivery platform show that our model achieves the highest F1 scores, with average improvements of 17.19\% in overall performance and 9.59\% on OOD transfer tests. These results demonstrate strong adaptability and robustness, highlighting the potential of our approach to enhance customer satisfaction and business value. Our work provides a robust and scalable solution for industrial abnormal event detection and response.

\section*{Limitations}

We have validated our approach on different model series, including DeepSeek-R1-Distill-Qwen-14B and Qwen-14B-Instruct, which meet the stringent latency and deployment requirements of our online business scenario. Although we have not systematically tested other model architectures or sizes, our results demonstrate practical applicability. Future work is encouraged to extend this evaluation to both larger and smaller models to further explore the scalability and robustness of our method.


\section*{Ethics Statement}

We have carefully considered the ethical implications of our research and provide the following statements:

\begin{itemize}
    \item Throughout this study, we have strictly followed established ethical guidelines, ensuring that our findings are reported honestly, transparently, and with full accuracy.
    \item No sensitive or confidential information was used at any stage of our research. All data and materials utilized are suitable for public release.
    \item The datasets employed in our experiments originate from publicly available and peer-reviewed scientific sources, supporting the transparency and reproducibility of our work.
    \item We offer detailed descriptions of the datasets and the hyper-parameter configurations used in our experiments to ensure the reproducibility and clarity of our results.
    \item In the interest of openness and to support future research, we have made our code available anonymously on GitHub and will fully open source it following the acceptance of our paper.
\end{itemize}


\bibliography{custom}

\begin{thebibliography}{63}
\providecommand{\natexlab}[1]{#1}

\bibitem[{Ahmadian et~al.(2024)Ahmadian, Cremer, Gall{\'e}, Fadaee, Kreutzer, Pietquin, {\"U}st{\"u}n, and Hooker}]{ahmadian2024rloo}
Arash Ahmadian, Chris Cremer, Matthias Gall{\'e}, Marzieh Fadaee, Julia Kreutzer, Olivier Pietquin, Ahmet {\"U}st{\"u}n, and Sara Hooker. 2024.
\newblock Back to basics: Revisiting reinforce style optimization for learning from human feedback in llms.
\newblock \emph{arXiv preprint arXiv:2402.14740}.

\bibitem[{Arrieta et~al.(2025)Arrieta, Ugarte, Valle, Parejo, and Segura}]{arrieta2025o3}
Aitor Arrieta, Miriam Ugarte, Pablo Valle, Jos{\'e}~Antonio Parejo, and Sergio Segura. 2025.
\newblock o3-mini vs deepseek-r1: Which one is safer?
\newblock \emph{arXiv preprint arXiv:2501.18438}.

\bibitem[{Bai et~al.(2022)Bai, Jones, Ndousse, Askell, Chen, DasSarma, Drain, Fort, Ganguli, Henighan et~al.}]{bai2022training}
Yuntao Bai, Andy Jones, Kamal Ndousse, Amanda Askell, Anna Chen, Nova DasSarma, Dawn Drain, Stanislav Fort, Deep Ganguli, Tom Henighan, and 1 others. 2022.
\newblock Training a helpful and harmless assistant with reinforcement learning from human feedback.
\newblock \emph{arXiv preprint arXiv:2204.05862}.

\bibitem[{Baxter(1995)}]{baxter1995markov}
Laurence~A Baxter. 1995.
\newblock Markov decision processes: Discrete stochastic dynamic programming.

\bibitem[{Chen et~al.(2024)Chen, Xiao, Zhang, Luo, Lian, and Liu}]{chen2024bge}
Jianlv Chen, Shitao Xiao, Peitian Zhang, Kun Luo, Defu Lian, and Zheng Liu. 2024.
\newblock Bge m3-embedding: Multi-lingual, multi-functionality, multi-granularity text embeddings through self-knowledge distillation.
\newblock \emph{arXiv preprint arXiv:2402.03216}.

\bibitem[{Chen et~al.(2025)Chen, Li, Wang, Jin, Qian, Wang, Wang, Zhang, Zhang, Zhang et~al.}]{chen2025rmr1}
Xiusi Chen, Gaotang Li, Ziqi Wang, Bowen Jin, Cheng Qian, Yu~Wang, Hongru Wang, Yu~Zhang, Denghui Zhang, Tong Zhang, and 1 others. 2025.
\newblock Rm-r1: Reward modeling as reasoning.
\newblock \emph{arXiv preprint arXiv:2505.02387}.

\bibitem[{Chu et~al.(2025)Chu, Zhai, Yang, Tong, Xie, Schuurmans, Le, Levine, and Ma}]{chu2025sft}
Tianzhe Chu, Yuexiang Zhai, Jihan Yang, Shengbang Tong, Saining Xie, Dale Schuurmans, Quoc~V Le, Sergey Levine, and Yi~Ma. 2025.
\newblock Sft memorizes, rl generalizes: A comparative study of foundation model post-training.
\newblock \emph{arXiv preprint arXiv:2501.17161}.

\bibitem[{Dai et~al.(2024)Dai, Lu, Zivlak, Lali{\'c}, and Lalic}]{dai2024does}
Tao Dai, Yidan Lu, Nikola Zivlak, Danijela~{\'C}iri{\'c} Lali{\'c}, and Bojan Lalic. 2024.
\newblock How does queueing information in pre-sales call centers affect customer repurchase behavior.
\newblock \emph{Asia-Pacific Journal of Operational Research}, (41 (4)):26--p.

\bibitem[{DeepMind(2024)}]{gemeni_thinking}
Google DeepMind. 2024.
\newblock Gemini 2.0 flash thinking.
\newblock \emph{URL https://deepmind.google/technologies/gemini/flash-thinking/}.

\bibitem[{Devlin et~al.(2019)Devlin, Chang, Lee, and Toutanova}]{devlin2019bert}
Jacob Devlin, Ming-Wei Chang, Kenton Lee, and Kristina Toutanova. 2019.
\newblock Bert: Pre-training of deep bidirectional transformers for language understanding.
\newblock In \emph{Proceedings of the 2019 conference of the North American chapter of the association for computational linguistics: human language technologies, volume 1 (long and short papers)}, pages 4171--4186.

\bibitem[{Evstafev(2025)}]{evstafev2025token}
Evgenii Evstafev. 2025.
\newblock Token-hungry, yet precise: Deepseek r1 highlights the need for multi-step reasoning over speed in math.
\newblock \emph{arXiv preprint arXiv:2501.18576}.

\bibitem[{Gao et~al.(2021)Gao, Zhang, Wu, Hu, Ru, Hao, He, and Sun}]{gao2021deep}
Chengliang Gao, Fan Zhang, Guanqun Wu, Qiwan Hu, Qiang Ru, Jinghua Hao, Renqing He, and Zhizhao Sun. 2021.
\newblock A deep learning method for route and time prediction in food delivery service.
\newblock In \emph{Proceedings of the 27th ACM SIGKDD Conference on Knowledge Discovery \& Data Mining}, pages 2879--2889.

\bibitem[{Guo et~al.(2025{\natexlab{a}})Guo, Yang, Zhang, Song, Zhang, Xu, Zhu, Ma, Wang, Bi et~al.}]{guo2025deepseek}
Daya Guo, Dejian Yang, Haowei Zhang, Junxiao Song, Ruoyu Zhang, Runxin Xu, Qihao Zhu, Shirong Ma, Peiyi Wang, Xiao Bi, and 1 others. 2025{\natexlab{a}}.
\newblock Deepseek-r1: Incentivizing reasoning capability in llms via reinforcement learning.
\newblock \emph{arXiv preprint arXiv:2501.12948}.

\bibitem[{Guo et~al.(2025{\natexlab{b}})Guo, Yang, Zhang, Song, Zhang, Xu, Zhu, Ma, Wang, Bi et~al.}]{guo2025deepseekr1}
Daya Guo, Dejian Yang, Haowei Zhang, Junxiao Song, Ruoyu Zhang, Runxin Xu, Qihao Zhu, Shirong Ma, Peiyi Wang, Xiao Bi, and 1 others. 2025{\natexlab{b}}.
\newblock Deepseek-r1: Incentivizing reasoning capability in llms via reinforcement learning.
\newblock \emph{arXiv preprint arXiv:2501.12948}.

\bibitem[{Howard(1960)}]{howard1960dynamic}
Ronald~A Howard. 1960.
\newblock Dynamic programming and markov processes.

\bibitem[{Hu(2025)}]{hu2025reinforce++}
Jian Hu. 2025.
\newblock Reinforce++: A simple and efficient approach for aligning large language models.
\newblock \emph{arXiv preprint arXiv:2501.03262}.

\bibitem[{Jaech et~al.(2024)Jaech, Kalai, Lerer, Richardson, El-Kishky, Low, Helyar, Madry, Beutel, Carney et~al.}]{jaech2024openai}
Aaron Jaech, Adam Kalai, Adam Lerer, Adam Richardson, Ahmed El-Kishky, Aiden Low, Alec Helyar, Aleksander Madry, Alex Beutel, Alex Carney, and 1 others. 2024.
\newblock Openai o1 system card.
\newblock \emph{arXiv preprint arXiv:2412.16720}.

\bibitem[{Kaplan et~al.(2020)Kaplan, McCandlish, Henighan, Brown, Chess, Child, Gray, Radford, Wu, and Amodei}]{kaplan2020scaling}
Jared Kaplan, Sam McCandlish, Tom Henighan, Tom~B Brown, Benjamin Chess, Rewon Child, Scott Gray, Alec Radford, Jeffrey Wu, and Dario Amodei. 2020.
\newblock Scaling laws for neural language models.
\newblock \emph{arXiv preprint arXiv:2001.08361}.

\bibitem[{Kwon et~al.(2023)Kwon, Li, Zhuang, Sheng, Zheng, Yu, Gonzalez, Zhang, and Stoica}]{kwon2023vllm}
Woosuk Kwon, Zhuohan Li, Siyuan Zhuang, Ying Sheng, Lianmin Zheng, Cody~Hao Yu, Joseph~E. Gonzalez, Hao Zhang, and Ion Stoica. 2023.
\newblock Efficient memory management for large language model serving with pagedattention.
\newblock In \emph{Proceedings of the ACM SIGOPS 29th Symposium on Operating Systems Principles}.

\bibitem[{Lai et~al.(2025)Lai, Zhong, Li, Zhao, and Yang}]{lai2025med}
Yuxiang Lai, Jike Zhong, Ming Li, Shitian Zhao, and Xiaofeng Yang. 2025.
\newblock Med-r1: Reinforcement learning for generalizable medical reasoning in vision-language models.
\newblock \emph{arXiv preprint arXiv:2503.13939}.

\bibitem[{Li et~al.(2023{\natexlab{a}})Li, Zhang, Zhang, Long, Xie, and Zhang}]{li2023gte}
Zehan Li, Xin Zhang, Yanzhao Zhang, Dingkun Long, Pengjun Xie, and Meishan Zhang. 2023{\natexlab{a}}.
\newblock Towards general text embeddings with multi-stage contrastive learning.
\newblock \emph{arXiv preprint arXiv:2308.03281}.

\bibitem[{Li et~al.(2023{\natexlab{b}})Li, Xu, Zhang, Lin, Yu, Sun, and Luo}]{li2023remax}
Ziniu Li, Tian Xu, Yushun Zhang, Zhihang Lin, Yang Yu, Ruoyu Sun, and Zhi-Quan Luo. 2023{\natexlab{b}}.
\newblock Remax: A simple, effective, and efficient reinforcement learning method for aligning large language models.
\newblock \emph{arXiv preprint arXiv:2310.10505}.

\bibitem[{Liu et~al.(2025{\natexlab{a}})Liu, Guo, Lou, Zeng, Niu, Wang, Xu, Cai, Yang, Zhao et~al.}]{liu2025fin}
Zhaowei Liu, Xin Guo, Fangqi Lou, Lingfeng Zeng, Jinyi Niu, Zixuan Wang, Jiajie Xu, Weige Cai, Ziwei Yang, Xueqian Zhao, and 1 others. 2025{\natexlab{a}}.
\newblock Fin-r1: A large language model for financial reasoning through reinforcement learning.
\newblock \emph{arXiv preprint arXiv:2503.16252}.

\bibitem[{Liu et~al.(2025{\natexlab{b}})Liu, Chen, Li, Qi, Pang, Du, Lee, and Lin}]{liu2025understanding}
Zichen Liu, Changyu Chen, Wenjun Li, Penghui Qi, Tianyu Pang, Chao Du, Wee~Sun Lee, and Min Lin. 2025{\natexlab{b}}.
\newblock Understanding r1-zero-like training: A critical perspective.
\newblock \emph{arXiv preprint arXiv:2503.20783}.

\bibitem[{Meng et~al.(2024)Meng, Xia, and Chen}]{meng2024simpo}
Yu~Meng, Mengzhou Xia, and Danqi Chen. 2024.
\newblock Simpo: Simple preference optimization with a reference-free reward.
\newblock \emph{Advances in Neural Information Processing Systems}, 37:124198--124235.

\bibitem[{Min et~al.(2022)Min, Lyu, Holtzman, Artetxe, Lewis, Hajishirzi, and Zettlemoyer}]{min2022rethinking}
Sewon Min, Xinxi Lyu, Ari Holtzman, Mikel Artetxe, Mike Lewis, Hannaneh Hajishirzi, and Luke Zettlemoyer. 2022.
\newblock Rethinking the role of demonstrations: What makes in-context learning work?
\newblock \emph{arXiv preprint arXiv:2202.12837}.

\bibitem[{Mondillo et~al.(2025)Mondillo, Colosimo, Perrotta, Frattolillo, and Masino}]{mondillo2025comparative}
Gianluca Mondillo, Simone Colosimo, Alessandra Perrotta, Vittoria Frattolillo, and Mariapia Masino. 2025.
\newblock Comparative evaluation of advanced ai reasoning models in pediatric clinical decision support: Chatgpt o1 vs. deepseek-r1.
\newblock \emph{medRxiv}, pages 2025--01.

\bibitem[{Naik et~al.(2024)Naik, Naik, and Naik}]{naik2024leveraging}
Dishita Naik, Ishita Naik, and Nitin Naik. 2024.
\newblock Leveraging the use of chatgpt: exploring its real-world applications including their related ethical and regulatory considerations.
\newblock In \emph{The International Conference on Computing, Communication, Cybersecurity \& AI}, pages 649--667. Springer.

\bibitem[{Ngai et~al.(2021)Ngai, Lee, Luo, Chan, and Liang}]{ngai2021intelligent}
Eric~WT Ngai, Maggie~CM Lee, Mei Luo, Patrick~SL Chan, and Tenglu Liang. 2021.
\newblock An intelligent knowledge-based chatbot for customer service.
\newblock \emph{Electronic Commerce Research and Applications}, 50:101098.

\bibitem[{OpenAI(2023)}]{openai2023gpt4}
OpenAI. 2023.
\newblock \href {https://arxiv.org/abs/2303.08774} {Gpt-4 technical report}.
\newblock \emph{Preprint}, arXiv:2303.08774.

\bibitem[{OpenAI(2024)}]{o12024openaio1}
OpenAI. 2024.
\newblock \href {https://openai.com/index/learning-to-reason-with-llms} {Learning to reason with language models}.

\bibitem[{Ouyang et~al.(2022)Ouyang, Wu, Jiang, Almeida, Wainwright, Mishkin, Zhang, Agarwal, Slama, Ray et~al.}]{ouyang2022training}
Long Ouyang, Jeffrey Wu, Xu~Jiang, Diogo Almeida, Carroll Wainwright, Pamela Mishkin, Chong Zhang, Sandhini Agarwal, Katarina Slama, Alex Ray, and 1 others. 2022.
\newblock Training language models to follow instructions with human feedback.
\newblock \emph{Advances in Neural Information Processing Systems}, 35:27730--27744.

\bibitem[{Prabhu et~al.(2021)Prabhu, Mohamed, and Misra}]{prabhu2021multi}
Sumanth Prabhu, Moosa Mohamed, and Hemant Misra. 2021.
\newblock Multi-class text classification using bert-based active learning.
\newblock \emph{arXiv preprint arXiv:2104.14289}.

\bibitem[{Qasim et~al.(2022)Qasim, Bangyal, Alqarni, and Ali~Almazroi}]{qasim2022fine}
Rukhma Qasim, Waqas~Haider Bangyal, Mohammed~A Alqarni, and Abdulwahab Ali~Almazroi. 2022.
\newblock A fine-tuned bert-based transfer learning approach for text classification.
\newblock \emph{Journal of healthcare engineering}, 2022(1):3498123.

\bibitem[{Qwen et~al.(2025)Qwen, :, Yang, Yang, Zhang, Hui, Zheng, Yu, Li, Liu, Huang, Wei, Lin, Yang, Tu, Zhang, Yang, Yang, Zhou, Lin, Dang, Lu, Bao, Yang, Yu, Li, Xue, Zhang, Zhu, Men, Lin, Li, Tang, Xia, Ren, Ren, Fan, Su, Zhang, Wan, Liu, Cui, Zhang, and Qiu}]{qwen2025qwen25technicalreport}
Qwen, :, An~Yang, Baosong Yang, Beichen Zhang, Binyuan Hui, Bo~Zheng, Bowen Yu, Chengyuan Li, Dayiheng Liu, Fei Huang, Haoran Wei, Huan Lin, Jian Yang, Jianhong Tu, Jianwei Zhang, Jianxin Yang, Jiaxi Yang, Jingren Zhou, and 25 others. 2025.
\newblock \href {https://arxiv.org/abs/2412.15115} {Qwen2.5 technical report}.
\newblock \emph{Preprint}, arXiv:2412.15115.

\bibitem[{Rafailov et~al.(2023)Rafailov, Sharma, Mitchell, Manning, Ermon, and Finn}]{rafailov2023direct}
Rafael Rafailov, Archit Sharma, Eric Mitchell, Christopher~D Manning, Stefano Ermon, and Chelsea Finn. 2023.
\newblock Direct preference optimization: Your language model is secretly a reward model.
\newblock \emph{Advances in Neural Information Processing Systems}, 36:53728--53741.

\bibitem[{Ravikumar et~al.(2025)Ravikumar, Cabral, AlSubhi, and Puthukulam}]{ravikumar2025study}
Anitha Ravikumar, Donalie~H Cabral, Sheikha~Said AlSubhi, and Gopalan Puthukulam. 2025.
\newblock A study on the application of ai chatbots for customized product recommendations and uninterrupted order fulfillment for enhanced customer experience using kano model.
\newblock In \emph{Innovative and Intelligent Digital Technologies; Towards an Increased Efficiency: Volume 2}, pages 403--415. Springer.

\bibitem[{Schaul et~al.(2019)Schaul, Borsa, Modayil, and Pascanu}]{schaul2019ray}
Tom Schaul, Diana Borsa, Joseph Modayil, and Razvan Pascanu. 2019.
\newblock Ray interference: a source of plateaus in deep reinforcement learning.
\newblock \emph{arXiv preprint arXiv:1904.11455}.

\bibitem[{Schulman et~al.(2017)Schulman, Wolski, Dhariwal, Radford, and Klimov}]{schulman2017proximal}
John Schulman, Filip Wolski, Prafulla Dhariwal, Alec Radford, and Oleg Klimov. 2017.
\newblock Proximal policy optimization algorithms.
\newblock \emph{arXiv preprint arXiv:1707.06347}.

\bibitem[{Shao et~al.(2024{\natexlab{a}})Shao, Wang, Zhu, Xu, Song, Bi, Zhang, Zhang, Li, Wu, and Guo}]{Shao24:Deepseekmath}
Zhihong Shao, Peiyi Wang, Qihao Zhu, Runxin Xu, Junxiao Song, Xiao Bi, Haowei Zhang, Mingchuan Zhang, Y.~K. Li, Y.~Wu, and Daya Guo. 2024{\natexlab{a}}.
\newblock \href {https://arxiv.org/abs/2402.03300} {Deepseekmath: Pushing the limits of mathematical reasoning in open language models}.
\newblock \emph{Preprint}, arXiv:2402.03300.

\bibitem[{Shao et~al.(2024{\natexlab{b}})Shao, Wang, Zhu, Xu, Song, Bi, Zhang, Zhang, Li, Wu et~al.}]{shao2024deepseekmath}
Zhihong Shao, Peiyi Wang, Qihao Zhu, Runxin Xu, Junxiao Song, Xiao Bi, Haowei Zhang, Mingchuan Zhang, YK~Li, Y~Wu, and 1 others. 2024{\natexlab{b}}.
\newblock Deepseekmath: Pushing the limits of mathematical reasoning in open language models.
\newblock \emph{arXiv preprint arXiv:2402.03300}.

\bibitem[{Sheng et~al.(2024)Sheng, Zhang, Ye, Wu, Zhang, Zhang, Peng, Lin, and Wu}]{sheng2024verl}
Guangming Sheng, Chi Zhang, Zilingfeng Ye, Xibin Wu, Wang Zhang, Ru~Zhang, Yanghua Peng, Haibin Lin, and Chuan Wu. 2024.
\newblock Hybridflow: A flexible and efficient rlhf framework.
\newblock \emph{arXiv preprint arXiv: 2409.19256}.

\bibitem[{Shoeybi et~al.(2019)Shoeybi, Patwary, Puri, LeGresley, Casper, and Catanzaro}]{shoeybi2019megatron}
Mohammad Shoeybi, Mostofa Patwary, Raul Puri, Patrick LeGresley, Jared Casper, and Bryan Catanzaro. 2019.
\newblock Megatron-lm: Training multi-billion parameter language models using model parallelism.
\newblock \emph{arXiv preprint arXiv:1909.08053}.

\bibitem[{Slade and Gedeon(1993)}]{slade1993bimodal}
P~Slade and Tam{\'a}s~D Gedeon. 1993.
\newblock Bimodal distribution removal.
\newblock In \emph{International Workshop on Artificial Neural Networks}, pages 249--254. Springer.

\bibitem[{Snell et~al.(2024)Snell, Lee, Xu, and Kumar}]{snell2024scaling}
Charlie Snell, Jaehoon Lee, Kelvin Xu, and Aviral Kumar. 2024.
\newblock Scaling llm test-time compute optimally can be more effective than scaling model parameters.
\newblock \emph{arXiv preprint arXiv:2408.03314}.

\bibitem[{Sun and Moens(2023)}]{sun2023fine}
Jingyuan Sun and Marie-Francine Moens. 2023.
\newblock Fine-tuned vs. prompt-tuned supervised representations: Which better account for brain language representations?
\newblock \emph{arXiv preprint arXiv:2310.01854}.

\bibitem[{Team et~al.(2025)Team, Du, Gao, Xing, Jiang, Chen, Li, Xiao, Du, Liao et~al.}]{team2025kimi}
Kimi Team, Angang Du, Bofei Gao, Bowei Xing, Changjiu Jiang, Cheng Chen, Cheng Li, Chenjun Xiao, Chenzhuang Du, Chonghua Liao, and 1 others. 2025.
\newblock Kimi k1. 5: Scaling reinforcement learning with llms.
\newblock \emph{arXiv preprint arXiv:2501.12599}.

\bibitem[{Touvron et~al.(2023)Touvron, Lavril, Izacard, Martinet, Lachaux, Lacroix, Rozi{\`e}re, Goyal, Hambro, Azhar et~al.}]{touvron2023llama}
Hugo Touvron, Thibaut Lavril, Gautier Izacard, Xavier Martinet, Marie-Anne Lachaux, Timoth{\'e}e Lacroix, Baptiste Rozi{\`e}re, Naman Goyal, Eric Hambro, Faisal Azhar, and 1 others. 2023.
\newblock Llama: Open and efficient foundation language models.
\newblock \emph{arXiv preprint arXiv:2302.13971}.

\bibitem[{Wake et~al.(2023)Wake, Kanehira, Sasabuchi, Takamatsu, and Ikeuchi}]{wake2023chatgpt}
Naoki Wake, Atsushi Kanehira, Kazuhiro Sasabuchi, Jun Takamatsu, and Katsushi Ikeuchi. 2023.
\newblock Chatgpt empowered long-step robot control in various environments: A case application.
\newblock \emph{arXiv preprint arXiv:2304.03893}.

\bibitem[{Wei et~al.(2021)Wei, Bosma, Zhao, Guu, Yu, Lester, Du, Dai, and Le}]{wei2021finetuned}
Jason Wei, Maarten Bosma, Vincent~Y Zhao, Kelvin Guu, Adams~Wei Yu, Brian Lester, Nan Du, Andrew~M Dai, and Quoc~V Le. 2021.
\newblock Finetuned language models are zero-shot learners.
\newblock \emph{arXiv preprint arXiv:2109.01652}.

\bibitem[{Wei et~al.(2022{\natexlab{a}})Wei, Wang, Schuurmans, Bosma, Xia, Chi, Le, Zhou et~al.}]{wei2022chain}
Jason Wei, Xuezhi Wang, Dale Schuurmans, Maarten Bosma, Fei Xia, Ed~Chi, Quoc~V Le, Denny Zhou, and 1 others. 2022{\natexlab{a}}.
\newblock Chain-of-thought prompting elicits reasoning in large language models.
\newblock \emph{Advances in neural information processing systems}, 35:24824--24837.

\bibitem[{Wei et~al.(2022{\natexlab{b}})Wei, Wang, Schuurmans, Bosma, Xia, Chi, Le, Zhou et~al.}]{wei2022cot}
Jason Wei, Xuezhi Wang, Dale Schuurmans, Maarten Bosma, Fei Xia, Ed~Chi, Quoc~V Le, Denny Zhou, and 1 others. 2022{\natexlab{b}}.
\newblock Chain-of-thought prompting elicits reasoning in large language models.
\newblock \emph{Advances in neural information processing systems}, 35:24824--24837.

\bibitem[{Xie et~al.(2025)Xie, Gao, Ren, Luo, Hong, Dai, Zhou, Qiu, Wu, and Luo}]{Xie25:logicrl}
Tian Xie, Zitian Gao, Qingnan Ren, Haoming Luo, Yuqian Hong, Bryan Dai, Joey Zhou, Kai Qiu, Zhirong Wu, and Chong Luo. 2025.
\newblock \href {https://arxiv.org/abs/2502.14768} {Logic-rl: Unleashing llm reasoning with rule-based reinforcement learning}.
\newblock \emph{Preprint}, arXiv:2502.14768.

\bibitem[{Yang et~al.(2024)Yang, Lau, and Wang}]{yang2024optimization}
Jingfeng Yang, Hoong~Chuin Lau, and Hai Wang. 2024.
\newblock Optimization of customer service and driver dispatch areas for on-demand food delivery.
\newblock \emph{Transportation Research Part C: Emerging Technologies}, 165:104653.

\bibitem[{Yao et~al.(2023)Yao, Yu, Zhao, Shafran, Griffiths, Cao, and Narasimhan}]{yao2023tree}
Shunyu Yao, Dian Yu, Jeffrey Zhao, Izhak Shafran, Tom Griffiths, Yuan Cao, and Karthik Narasimhan. 2023.
\newblock Tree of thoughts: Deliberate problem solving with large language models.
\newblock \emph{Advances in neural information processing systems}, 36:11809--11822.

\bibitem[{Yu et~al.(2025)Yu, Zhang, Zhu, Yuan, Zuo, Yue, Dai, Fan, Liu, Liu et~al.}]{yu2025dapo}
Qiying Yu, Zheng Zhang, Ruofei Zhu, Yufeng Yuan, Xiaochen Zuo, Yu~Yue, Weinan Dai, Tiantian Fan, Gaohong Liu, Lingjun Liu, and 1 others. 2025.
\newblock Dapo: An open-source llm reinforcement learning system at scale.
\newblock \emph{arXiv preprint arXiv:2503.14476}.

\bibitem[{Yu et~al.(2019)Yu, Su, and Luo}]{yu2019improving}
Shanshan Yu, Jindian Su, and Da~Luo. 2019.
\newblock Improving bert-based text classification with auxiliary sentence and domain knowledge.
\newblock \emph{IEEE Access}, 7:176600--176612.

\bibitem[{Yue et~al.(2025)Yue, Yuan, Yu, Zuo, Zhu, Xu, Chen, Wang, Fan, Du et~al.}]{yue2025vapo}
Yu~Yue, Yufeng Yuan, Qiying Yu, Xiaochen Zuo, Ruofei Zhu, Wenyuan Xu, Jiaze Chen, Chengyi Wang, TianTian Fan, Zhengyin Du, and 1 others. 2025.
\newblock Vapo: Efficient and reliable reinforcement learning for advanced reasoning tasks.
\newblock \emph{arXiv preprint arXiv:2504.05118}.

\bibitem[{Zhang et~al.(2025)Zhang, Deng, Lin, Wang, Ng, Ye, Li, Xiao, Mo, Zhang et~al.}]{zhang2025100}
Chong Zhang, Yue Deng, Xiang Lin, Bin Wang, Dianwen Ng, Hai Ye, Xingxuan Li, Yao Xiao, Zhanfeng Mo, Qi~Zhang, and 1 others. 2025.
\newblock 100 days after deepseek-r1: A survey on replication studies and more directions for reasoning language models.
\newblock \emph{arXiv preprint arXiv:2505.00551}.

\bibitem[{Zhang and Zuo(2025)}]{zhang2025grpolead}
Jixiao Zhang and Chunsheng Zuo. 2025.
\newblock Grpo-lead: A difficulty-aware reinforcement learning approach for concise mathematical reasoning in language models.
\newblock \emph{arXiv preprint arXiv:2504.09696}.

\bibitem[{Zhou et~al.(2022)Zhou, Sch{\"a}rli, Hou, Wei, Scales, Wang, Schuurmans, Cui, Bousquet, Le et~al.}]{zhou2022least}
Denny Zhou, Nathanael Sch{\"a}rli, Le~Hou, Jason Wei, Nathan Scales, Xuezhi Wang, Dale Schuurmans, Claire Cui, Olivier Bousquet, Quoc Le, and 1 others. 2022.
\newblock Least-to-most prompting enables complex reasoning in large language models.
\newblock \emph{arXiv preprint arXiv:2205.10625}.

\bibitem[{Zhu et~al.(2025)Zhu, Huang, Lyu, Zhang, Li, Shi, Wu, Mu, Wang, Zhao et~al.}]{zhu2025codev}
Yaoyu Zhu, Di~Huang, Hanqi Lyu, Xiaoyun Zhang, Chongxiao Li, Wenxuan Shi, Yutong Wu, Jianan Mu, Jinghua Wang, Yang Zhao, and 1 others. 2025.
\newblock Codev-r1: Reasoning-enhanced verilog generation.
\newblock \emph{arXiv preprint arXiv:2505.24183}.

\bibitem[{Zou et~al.(2021)Zou, Zhao, Kang, Lin, Peng, Jiang, Sun, Zhang, Huang, and Liu}]{zou2021topic}
Yicheng Zou, Lujun Zhao, Yangyang Kang, Jun Lin, Minlong Peng, Zhuoren Jiang, Changlong Sun, Qi~Zhang, Xuanjing Huang, and Xiaozhong Liu. 2021.
\newblock Topic-oriented spoken dialogue summarization for customer service with saliency-aware topic modeling.
\newblock In \emph{Proceedings of the AAAI Conference on Artificial Intelligence}, volume~35, pages 14665--14673.

\end{thebibliography}

\clearpage

\appendix

\onecolumn

\section{Related Work}

\paragraph{Large Reasoning Models.}
Large Reasoning Models (LRMs) enhance large language models by increasing inference-time computation~\citep{snell2024scaling} rather than simply scaling parameters.
Chain-of-Thought (CoT) prompting~\citep{wei2022cot,yao2023tree,zhou2022least}, which introduces intermediate reasoning steps, significantly boosts performance on complex tasks.
Building upon this, recent works further optimize reasoning via reinforcement learning, leading to advanced models such as OpenAI o1~\citep{o12024openaio1}, DeepSeek-R1~\citep{guo2025deepseekr1}, Kimi k1.5~\citep{team2025kimi}, and QwQ~\citep{qwen2025qwen25technicalreport}.
Trained with answer-based rewards, these models autonomously extend reasoning chains at inference, achieving substantial gains on challenging tasks like advanced mathematics and logical reasoning~\citep{zhang2025100,shao2024deepseekmath}.

\paragraph{Reinforcement Learning for LLM.}
Reinforcement Learning (RL) has become an essential tool for enhancing the reasoning capabilities and alignment of large language models (LLMs). Early research employed Proximal Policy Optimization (PPO)~\citep{schulman2017proximal} to fine-tune LLMs, thereby improving their performance across various natural language processing tasks by optimizing them according to human feedback, known as Reinforcement Learning from Human Feedback (RLHF)~\citep{bai2022training}. Building upon RLHF, several reinforcement learning preference optimization methods tailored for large language models have been developed, including DPO~\citep{rafailov2023direct}, SimPO~\citep{meng2024simpo}, ReMAX~\citep{li2023remax}, Reinforce++~\citep{hu2025reinforce++}, and RLOO~\citep{ahmadian2024rloo}. These approaches have effectively reduced the cost associated with preference optimization for large models and enhanced the stability of reinforcement learning processes for such models.

Following the demonstration of significant potential by large inference models such as OpenAI o1~\citep{o12024openaio1}, DeepSeek-R1~\citep{guo2025deepseekr1}, Kimi k1.5~\citep{team2025kimi}, and QwQ~\citep{qwen2025qwen25technicalreport}, the learning paradigm of Reinforcement Learning with Verifiable Reward (RLVR) has garnered increasing attention. Reinforcement learning techniques like GRPO~\citep{shao2024deepseekmath}, along with its improved variants such as Dr.GRPO~\citep{liu2025understanding}, DAPO~\citep{yu2025dapo}, and VAPO~\citep{yue2025vapo}, have been progressively applied. Additionally, more reinforcement learning models targeting vertical domains have emerged~\citep{Xie25:logicrl,zhu2025codev,liu2025fin,lai2025med}. Our approach centers on refining GRPO and DAPO and, for the first time, applying reinforcement learning techniques to event monitoring tasks within real-world business scenarios. The framework we propose is poised to establish a new benchmark for the application of large inference models in industry, further advancing operational efficiency and economic benefits in commercial settings.

\section{Implementation Details}
\label{app:trainingsetup}

\subsection{Training Setup}
We train all the methods with 32 A100-80GB SXM GPUs. For supervised fine-tuning (SFT), we utilize the Megatron-LM framework~\citep{shoeybi2019megatron}, which is a distributed training system designed for large-scale transformer models. Megatron-LM supports both tensor and pipeline parallelism, enabling efficient training of models with billions of parameters across multiple GPUs and nodes. This framework allows us to fully leverage the computational resources of our GPU cluster and achieve high training throughput and scalability. During the reinforcement learning stage, we adopt the VeRL framework~\citep{sheng2024verl}. VeRL is specifically designed for post-training large language models with reinforcement learning algorithms such as RLHF. It provides seamless integration with existing LLM infrastructures, including PyTorch FSDP, Megatron-LM, and vLLM. VeRL enables flexible and efficient RL training, supports modular APIs, and allows for easy extension to other training and inference frameworks. For inference, we employ the vLLM engine~\citep{kwon2023vllm}, which is an efficient and scalable LLM inference framework. vLLM supports asynchronous batch processing, distributed inference, and is optimized for high throughput and low latency on large language models. Its compatibility with both training and RL frameworks ensures a streamlined workflow from model training to deployment.

\subsection{Hyperparameters}

Here we summarize the training configurations for all compared methods, including our approach. The table provided summarizes the training configurations for various methods. Each method has specific hyperparameters tailored to optimize performance and efficiency during training and evaluation. These configurations are crucial for replicating the experiments and ensuring consistent results across different trials.

\begin{table}[htbp]
    \centering
    \renewcommand{\arraystretch}{1.2}
    \resizebox{0.96\textwidth}{!}{
    \begin{tabular}{lcccccccc}
        \toprule
        Method & data\_train\_batch\_size & ppo\_mini\_batch\_size & kl & length & lr & epoch & eval\_step & Others \\
        \midrule
        SFT & 512 & -- & -- & 8k & 1e-6 & 3 & 40 & -- \\
        GRPO & 256 & 128 & 0.001 & 8k & 1e-6 & 5 & 20 & -- \\
        DAPO & 256 & 128 & 0.0 & 8k & 1e-6 & 5 & 20 & -- \\
        APARL         & 256 & 128 & 0.001 & 8k & 1e-6 & 5 & 20 & $t=0.1$ \\
        \bottomrule
    \end{tabular}
    }
    \caption{Training configurations for different methods.}
    \label{tab:train_config}
\end{table}

\section{Supplementary Experimental Results}

\label{app:results}

\subsection{OOD Testing Detailed Results}

Table \ref{tab:domain_study} provides a comprehensive breakdown of the OOD testing results across three anonymized business domains. Each domain represents a distinct test set, with sizes of 8.8k, 8.5k, and 5k respectively. The table highlights the performance metrics (Precision, Recall, F1) for various models and training strategies, offering a detailed view beyond the average results presented in the main text.

Overall, the detailed results reaffirm the superiority of our APARL method, which consistently delivers higher precision and recall, translating into improved F1 scores across diverse OOD tasks. This robustness underlines the practical applicability of APARL in real-world scenarios where models face varied and unpredictable data distributions.

\begin{table}[ht]
\caption{Comparison across Different Domains and Training Strategies}
\label{tab:domain_study}
\centering
\resizebox{0.96\textwidth}{!}{ 
\begin{tabular}{llccc|ccc|ccc}
\toprule
\multirow{2}{*}{\textbf{Model}} & 
\multirow{2}{*}{\textbf{Method}} & 
\multicolumn{3}{c|}{\textbf{Anonymous Task 1}} & 
\multicolumn{3}{c|}{\textbf{Anonymous Task 2}} & 
\multicolumn{3}{c}{\textbf{Anonymous Task 3}} \\
\cmidrule(lr){3-5} \cmidrule(lr){6-8} \cmidrule(lr){9-11}
 & & P & R & F1 & P & R & F1 & P & R & F1 \\
\midrule

\multirow{1}{*}{\textbf{Small Models}} 
& \makecell[l]{BGE-M3\footnotemark} & / & / & / & / & / & / & / & / & / \\
\midrule

\multirow{3}{*}{\textbf{API Models}} 
& \makecell[l]{GPT-4o} & 70.84\% & 70.50\% & 70.67\% & 72.37\% & 69.60\% & 70.94\% & 72.24\% & 70.57\% & 71.40\% \\
& DeepSeek-V3 & 64.45\% & 70.35\% & 67.00\% & 69.40\% & 71.81\% & 70.50\% & 66.49\% & 72.93\% & 70.00\% \\
& DeepSeek-R1 & 69.45\% & 70.80\% & 69.80\% & 71.79\% & 72.43\% & 72.09\% & 70.05\% & 73.99\% & 71.90\% \\
\midrule

\multirow{5}{*}{\textbf{DeepSeek-R1-Distill-Qwen-14B}} 
& Base model & 52.16\% & \textbf{88.54\%} & 65.65\% & 56.34\% & \textbf{86.51\%} & 68.24\% & 55.57\% & \textbf{88.21\%} & 68.19\% \\
& +SFT & 66.55\% & 76.62\% & 71.10\% & 64.75\% & 80.69\% & 71.90\% & 67.09\% & 76.78\% & 71.60\% \\
& +GRPO & 66.68\% & 84.29\% & 74.46\% & 70.24\% & 80.79\% & 75.15\% & 67.99\% & 83.28\% & 74.86\% \\
& +DAPO & 74.57\% & 75.57\% & 75.07\% & 78.08\% & 71.60\% & 74.70\% & 75.74\% & 73.82\% & 74.77\% \\
& +Our APARL & \textbf{77.49\%} & 81.33\% & \textbf{79.36\%} & \textbf{81.88\%} & 76.85\% & \textbf{79.29\%} & \textbf{77.62\%} & 79.16\% & \textbf{78.38\%} \\
\midrule

\multirow{5}{*}{\textbf{Qwen-14B-Instruct}} 
& Base model & 54.28\% & \underline{87.49\%} & 66.99\% & 57.46\% & \underline{84.72\%} & 68.48\% & 58.74\% & \underline{86.36\%} & 69.92\% \\
& +SFT & 66.96\% & 79.42\% & 72.70\% & 64.98\% & 79.73\% & 71.50\% & 67.70\% & 78.33\% & 72.60\% \\
& +GRPO & \underline{77.27\%} & 73.63\% & 75.40\% & \underline{79.68\%} & 70.55\% & 74.80\% & \underline{77.27\%} & 69.83\% & 73.40\% \\
& +DAPO & 76.24\% & 78.77\% & 77.48\% & 79.31\% & 76.06\% & 77.65\% & 75.85\% & 77.72\% & 76.78\% \\
& +Our APARL & 76.80\% & 81.27\% & \underline{78.97\%} & 79.10\% & 79.27\% & \underline{79.18\%} & 76.46\% & 79.23\% & \underline{77.82\%} \\
\bottomrule
\end{tabular}
}
\end{table}

\footnotetext{The training and inference of BGE-M3 are based on fixed categories, making it unsuitable for direct transfer to out-of-domain tasks.}

\section{Case Study}

This case study demonstrates the application of an abnormal event detection system in the context of food delivery customer service interactions. By providing a detailed prompt containing event definitions and a real-world conversation between a customer and an agent, the system is able to accurately identify all relevant abnormal events present in the dialogue. After applying our APARL, the model's output becomes more precise and aligned with human reasoning.

\label{app:case_study}

\begin{figure*}
    \centering
    \includegraphics[width=0.8\linewidth]{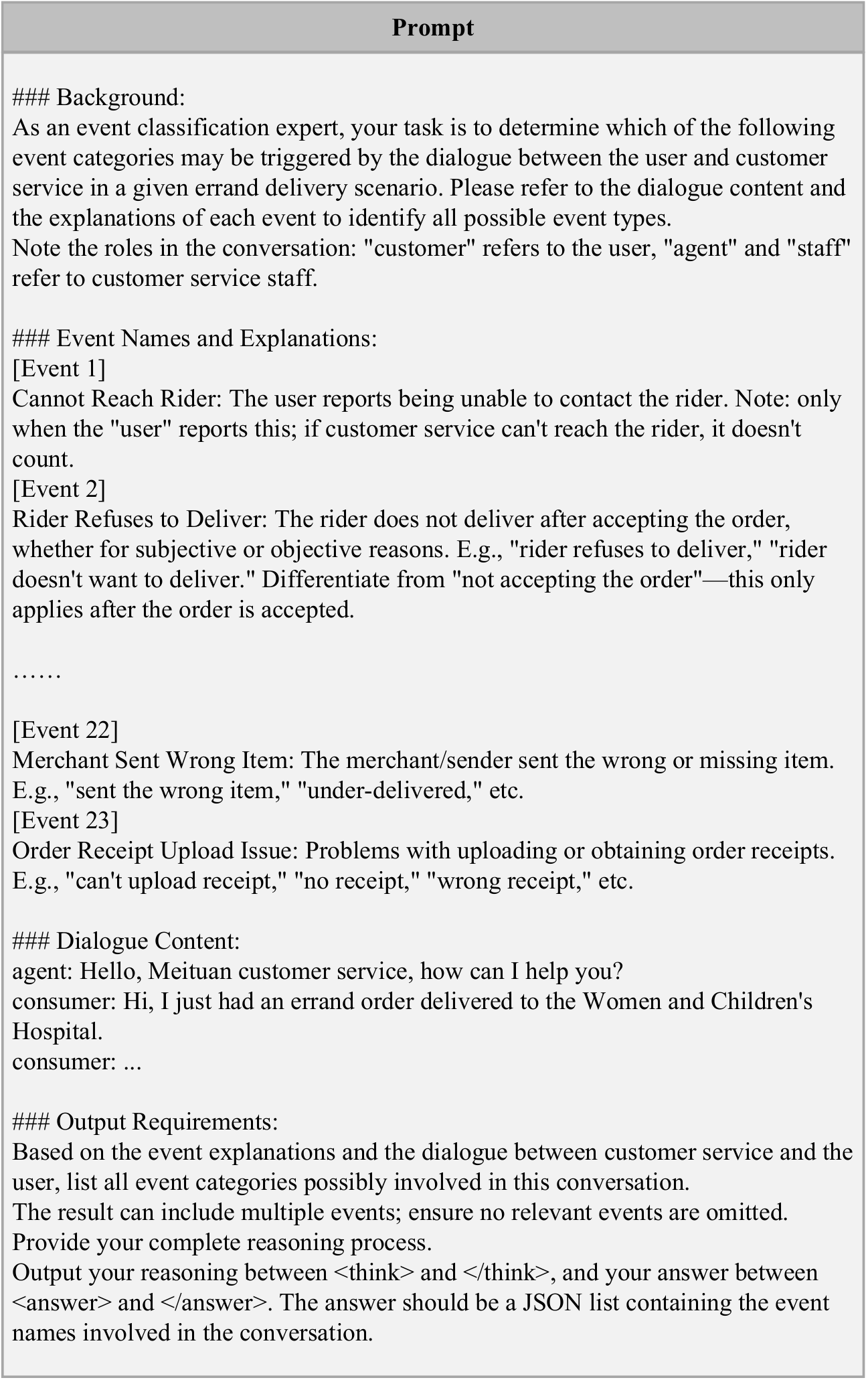}
    \caption{The prompt used for abnormal event detection in food delivery tasks.}
    \label{fig:prompt}
\end{figure*}

\begin{figure*}
    \centering
    \includegraphics[width=0.8\linewidth]{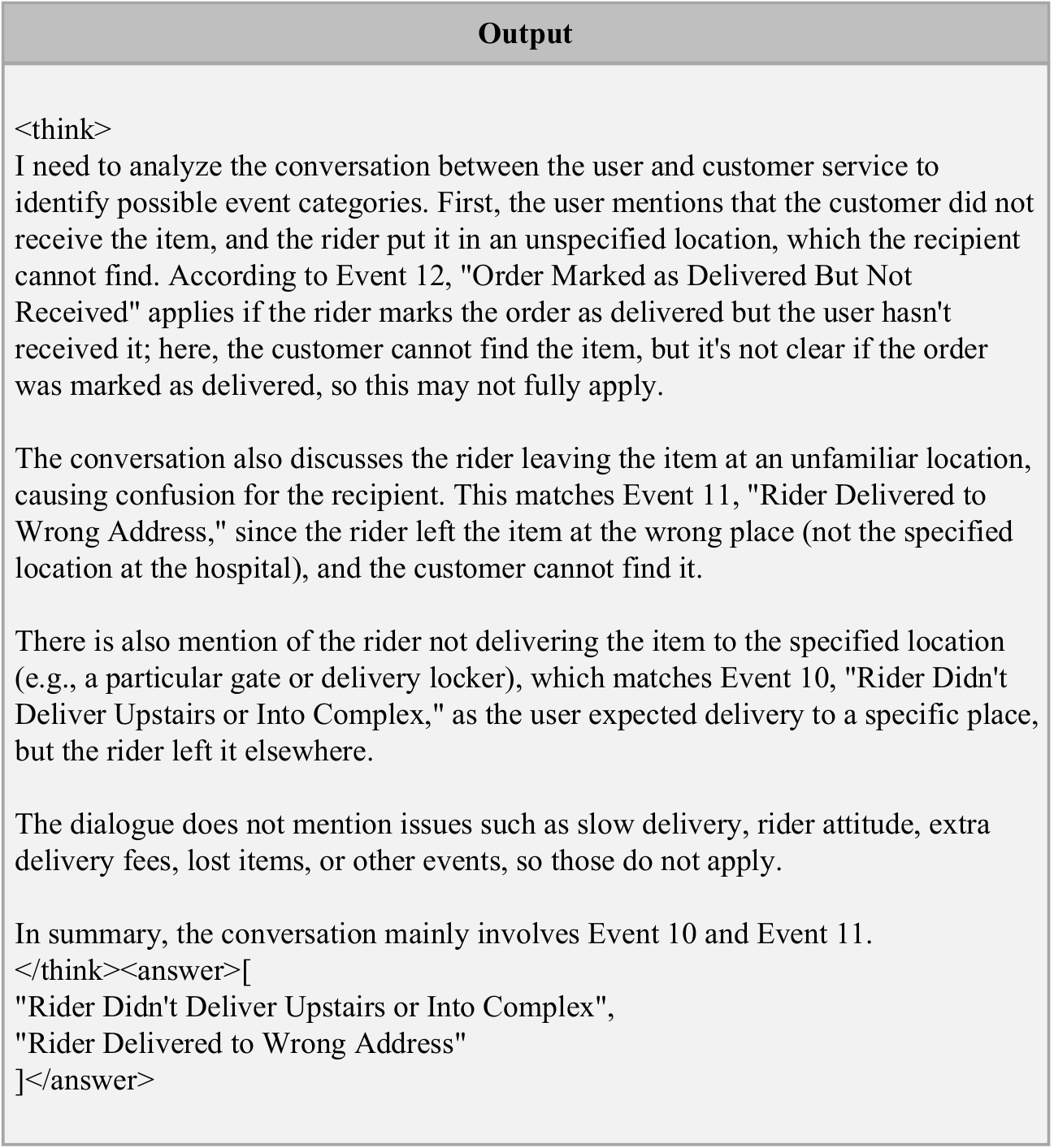}
    \caption{The output generated after ASRR reinforcement learning.}
    \label{fig:target}
\end{figure*}

The practical significance of this task lies in its ability to help food delivery platforms automatically and efficiently identify, categorize, and respond to various service exceptions in real time. This not only improves operational efficiency and user satisfaction by enabling faster resolution of issues, but also provides valuable insights for service quality monitoring, rider management, and process optimization. Ultimately, such intelligent event detection systems contribute to a safer, more reliable, and user-friendly food delivery ecosystem.

\end{document}